\documentclass[a4paper,review,11pt,authoryear]{elsarticle}
\usepackage{datetime2}
\usepackage{xcolor,bm} % The color
\usepackage{enumerate,enumitem} % the item list
\usepackage{booktabs,multirow,IEEEtrantools} % The table
\usepackage{amsmath,amsfonts,amsthm,amssymb, mathtools} %math
\usepackage{subcaption} % The figure
\usepackage{hyperref} % the ref
\usepackage[toc,page]{appendix} % the appendix
\usepackage[a4paper, margin=1in]{geometry}
\usepackage{color}
\usepackage{tikz} % the colored symbol

\newcommand{\beqa} {\begin{eqnarray}}
\newcommand{\eeqa} {\end{eqnarray}}

\newcommand{\n}{\nonumber\\}

\newcommand{\N}{{\mathbb{N}}}
\newcommand{\R}{{\mathbb{R}}}

\journal{}

\begin{document}
\date{}
\begin{frontmatter}
\title{Probabilistic Forecasting with Temporal Convolutional Neural Network}

\author[yitianaddress]{Yitian Chen} \ead{chenyitian@bigo.sg}
  \author[yanfeiaddress]{Yanfei Kang\corref{cor}} \ead{yanfeikang@buaa.edu.cn} 
  \cortext[cor]{Corresponding author}
  \author[yixiongaddress]{Yixiong Chen} \ead{chenyixiong516@msn.com}
  \author[zizhuoaddress]{Zizhuo Wang} \ead{wangzizhuo@cuhk.edu.cn}

  \address[yitianaddress]{Bigo Beijing R\&D Center, Bigo Inc., Beijing 100191, China.}
    \address[yanfeiaddress]{School of Economics and Management, Beihang
    University, Beijing 100191, China.}
  \address[yixiongaddress]{IBM China CIC, KIC Technology Center, Shanghai 200433, China.}
  \address[zizhuoaddress]{Institute for Data and Decision Analytics, The Chinese University of Hong Kong, Shenzhen, 518172, China.}

\begin{abstract}
We present a probabilistic forecasting framework based on convolutional neural network (CNN) for multiple related time series forecasting. The framework can be applied to estimate probability density under both parametric and non-parametric settings.
More specifically, stacked residual blocks based on dilated causal convolutional nets are constructed to capture the temporal dependencies of the series. Combined with representation learning, our approach is able to learn complex patterns such as seasonality, holiday effects within and across series, and to leverage those patterns for more accurate forecasts, especially when historical data is sparse or unavailable.  
Extensive empirical studies are performed on several real-world datasets, including datasets from JD.com, China's largest online retailer.
%The results show that our framework outperforms other state-of-the-art methods in both point and probabilistic forecasting.
The results show that our framework compares favorably to the state-of-the-art
in both point and probabilistic forecasting.
\end{abstract}
\end{frontmatter}

\section{Introduction}\label{introduction}
% Basic introduction
Time series forecasting plays a key role in many business decision-making scenarios, such as managing limited resources, optimizing operational processes, among others. Most existing forecasting methods focus on point forecasting, i.e., forecasting the conditional mean or median of future observations. However, probabilistic forecasting becomes increasingly important as it is able to extract richer information from historical data and better capture the uncertainty of the future. In retail business, probabilistic forecasting of product supply and demand is fundamental for successful procurement process and optimal inventory planning. 
Also, probabilistic shipment forecasting, i.e., generating probability distributions of the delivery volumes of packages, is the key component of the consequent logistics operations, such as labor resource planning and delivery vehicle deployment. 

In such circumstances, instead of predicting individual or a small number of time series, one needs to predict thousands or millions of related series. Moreover, there are many more challenges in real-world applications. For instance, new products emerge weekly on retail platforms and one often needs to forecast the demand of products without historical shopping festival data (e.g., Black Friday in North America, ``11.11'' shopping festival in China). Furthermore, forecasting often requires the consideration of exogenous variables that have significant influence on future demand (e.g., promotion plans provided by operations teams, accurate weather forecasts for brick and mortar retailers).
Such forecasting problems can be extended to a variety of domains. Examples include forecasting the web traffic for internet companies~\citep{Webtraffic}, the energy consumption for individual households, the load for servers in a data center~\citep{flunkert2017deepar} and traffic flows in transportation domain~\citep{lv2015traffic}.

% Classical time series forecasting
Classical forecasting methods, such as ARIMA~\citep[AutoRegressive Integrated Moving Average,][]{box2015time} and exponential smoothing~\citep{hyndman2008forecasting}, are widely employed for univariate base-level forecasting.  
To incorporate exogenous covariates, several extensions of these methods have been proposed, such as ARIMAX (AutoRegressive Integrated Moving Average with Explanatory Variable) and dynamic regression models~\citep{hyndman2018forecasting}.
These models are well-suited for applications in which the structure of the data is well understood and there is sufficient historical data.
However, working with thousands or millions of series requires prohibitive labor and computing resources for parameter estimation. 
Moreover, they are not applicable in situations where historical data is sparse or unavailable.

% The deep learning approach, the successe of RNN and seq2seq
%Deep learning methods such as 
Models based on Recurrent neural network (RNN)~\citep{graves2013generating} and the sequence to sequence (Seq2Seq) framework~\citep{cho2014learning,sutskever2014sequence} have achieved great success in many different sequential tasks such as machine translation~\citep{sutskever2014sequence}, language modeling~\citep{mikolov2010recurrent} and recently time series forecasting~\citep{laptev2017time,wen2017multi,flunkert2017deepar,rangapuram2018deep,SAGHEER2019203,SHEN2019}. 
For example, in the forecasting competition community, the Seq2Seq model based on a gated recurrent unit (GRU)~\citep{cho2014learning} won the Kaggle web traffic forecasting competition~\citep{webtraffic1st}. 
A hybrid model that combines exponential smoothing method and RNN won the M4 forecasting competition, which consists of 100,000 series with different seasonal patterns~\citep{makridakis2018m4}. 
However, training with back propagation through time (BPTT) algorithm often hampers efficient computation. In addition, training RNN can be remarkably difficult~\citep{werbos1990backpropagation,pascanu2013difficulty}. Dilated causal convolutional architectures, e.g., Wavenet~\citep{van2016wavenet}, offers an alternative for modeling sequential data.
By stacking layers of dilated causal convolutional nets, receptive fields can be increased, and the long-term correlations can be captured without violating the temporal orders. In addition, in dilated causal convolutional architectures, the training process can be performed in parallel, which guarantees computation efficiency.

Most Seq2Seq frameworks or Wavenet~\citep{van2016wavenet} are autoregressive generative models that factorize the joint distribution as a product of the conditionals. 
In this setting, a one-step-ahead prediction approach is adopted, i.e., first a prediction is generated by using the past observations, and the generated result is then fed back as the ground truth to make further forecasts.
More recent research shows that non-autoregressive approaches or direct prediction strategy, predicting observations of all time steps directly, can achieve better performances~\citep{gu2017non,bai2018empirical,wen2017multi}. 
In particular, non-autoregressive models are more robust to mis-specification by avoiding error accumulation and thus yield better prediction accuracy. Moreover, training over all the prediction horizons can be parallelized.

Having reviewing all these challenges and developments, in this paper, we propose the Deep Temporal Convolutional Network (\textnormal{DeepTCN}), a non-autoregressive probabilistic forecasting framework for large collections of related time series. The main contributions of the paper are as follows:
%\textcolor{red}{In particular, we generalize the classical approaches of combining information from past observations and exogenous covariates to multiple time series setting.}
%\textcolor{red}{Analogous to ARIMAX, information from both past observations and future exogenous covariates are combined to do more accurate recasting.}
%Analogous to ARIMAX, the proposed framework can deal with exogenous variables. Furthermore, it is more applicable to multiple time series setting, and situations with no (or little) historical data.
\begin{itemize}
    \item We propose a CNN-based forecasting framework that provides both parametric and non-parametric approaches for probability density estimation.
    \item The framework, being able to learn latent correlation among series and handle complex real-world forecasting situations such as data sparsity and cold starts, shows high scalability and extensibility.
    \item The model is very flexible and can include exogenous covariates such as additional promotion plans or weather forecasts.
    \item Extensive empirical studies show our framework compares favorably to state-of-the-art methods in both point forecasting and probabilistic forecasting tasks.   
\end{itemize}

The rest of this paper is organized as follows. Section \ref{relatedWork} provides a brief review of related work on time series forecasting and deep learning methods for forecasting. In Section \ref{method}, we describe the proposed forecasting method, including the neural network architectures, the probabilistic forecasting framework,  and the input features. 
We demonstrate the superiority of the proposed approach via extensive experiments in Section \ref{experiment} and conclude the paper in Section \ref{conclusion}.

\section{Related Work}\label{relatedWork}
% classic forecasting methods
Earlier studies on time series forecasting are mostly based on statistical models, which are mainly generative models based on state space framework such as exponential smoothing, ARIMA models and several other extensions. For these methods, \citet{hyndman2008forecasting} and \citet{box2015time} provide a comprehensive overview in the context of univariate forecasting.

% deep Learning for forecasting
In recent years, large number of related series are emerging in the routine functioning of many companies. Not sharing information from other time series, traditional univariate forecasting methods fit a model for each individual time series, and thus cannot learn across similar time series. Moreover, numerous researchers have shown that pure machine learning methods could not outperform statistical approaches in forecasting individual time series, the reasons for which can be attributed to overfitting and non-stationarity \citep{bandara2020forecasting, makridakis2018statistical}.
Therefore, methods that can provide forecasting on multiple series jointly have received increasing attention in the last few years~\citep[e.g.,][]{yu2016temporal}.

Both RNNs and CNNs have been shown to be able to model complex nonlinear feature interactions and yield substantial forecasting performances, especially when many related time series are available~\citep{smyl2016, laptev2017time, wen2017multi, flunkert2017deepar, rangapuram2018deep}. For example, Long Short-Term Memory (LSTM), one type of RNN architecture, won the CIF2016 forecasting competition for monthly time series~\citep{CIF2016}. \citet{bianchi2017overview} compare a variety of RNNs in their performances in the Short Term Load Forecasting problem. \citet{borovykh2017conditional} investigate the application of CNNs to financial time series forecasting. 

% probabilistic forecasting
To better understand the uncertainty of the future, probabilistic forecasting with deep learning models has attracted increasing attention. 
DeepAR~\citep{flunkert2017deepar}, which trains an auto-regressive RNN model on a rich collection of similar time series, produces more accurate probabilistic forecasts on several real-world data sets. The deep state space models (DeepState), presented by \citet{rangapuram2018deep}, combine state space models with deep learning and can retain data efficiency and interpretability while learning the complex patterns from raw data. Under a similar scheme, \citet{maddix2018deep} propose the combination of deep neural networks and Gaussian Process. More recently, \citet{gasthaus2019probabilistic} propose SQF-RNN, a probabilistic framework to model conditional quantile functions with isotonic splines, which allows more flexible output distributions.

% non-autoregressive
Most of these probabilistic forecasting frameworks are autoregressive models, which use recursive strategy to generate multi-step forecasts. 
In neural machine translation, non-autoregressive translation (NAT) models have achieved significant speedup at the cost of slightly inferior accuracy compared to autoregressive translation models~\citep{gu2017non}. For example, \citet{bai2018empirical} propose a \textnormal{non-autoregressive} framework based on dilated causal convolution and the empirical study on multiple datasets shows that the framework outperforms generic recurrent architectures such as LSTMs and GRUs. In forecasting applications, non-autoregressive approaches have also been shown to be less biased and more robust. Recently, \citet{wen2017multi} present a multi-horizon quantile recurrent forecaster to combine sequential neural nets and quantile regression \citep{koenker1978regression}. By training on all time points at the same time, their framework can significantly improve the training stability and the forecasting performances of recurrent nets.

%Our method differs from the aforementioned approaches in the following ways. Firstly, instead of applying gating mechanism used in Wavenet~\citep{van2016wavenet},  residual blocks are applied to stabilize the training of the network and help achieve superior forecasting accuracy. 
%Inspired by the models such as ARIMAX, a novel decoder based on a variant of the residual neural network is designed to incorporate information from both past observations and exogenous covariates. Finally, our model enjoys the flexibility to embrace a variety of probability density estimation approaches. 
Our method differs from the aforementioned approaches in the following ways.
First, stacked dilated causal convolutional nets are constructed to represent the encoder and model the stochastic process of historical observations of series.
Instead of applying gating mechanism (e.g., in Wavenet~\citep{van2016wavenet}),
residual blocks are used for the dilated causal convolutional nets to extract information of historical observations and help achieve superior forecasting accuracy.
Second, inspired by the dynamic regression models~\citep{pankratz2012forecasting} such as ARIMAX, in the decoder part, a novel variant of the residual neural network is proposed to incorporate information from both past observations and exogenous covariates.
Finally, our model enjoys the flexibility to embrace a variety of probability density estimation approaches.  %yitian

\section{Method}\label{method}
%% The problem
A general probabilistic forecasting problem for multiple related time series can be described as follows:
Given a set of time series ${\mathbf y}_{1:t} = \{y_{1:t}^{(i)}\}_{i=1}^N$,  we denote the future time series as ${\mathbf y}_{(t+1):(t+\Omega)} = \{y_{(t+1):(t+\Omega)}^{(i)}\}_{i=1}^N$, where $N$ is the number of series, $t$ is the length of the historical observations and $\Omega$ is the length of the forecasting horizon. Our goal is to model the conditional distribution of the future time series $P\left({\mathbf y}_{(t+1):(t+\Omega)}|{{\mathbf y}_{1:t}}\right)$. 

Classical generative models are often used to model time series data, which factorize the joint probability of future observations given the past information as the product of conditional probabilities:
\begin{equation}\label{cd1}
    P\left({\mathbf y}_{(t+1):(t+\Omega)}|{\mathbf y}_{1:t}\right) = \prod\limits_{\omega=1}^{\Omega} p({\mathbf y}_{t+\omega}|{\mathbf y}_{1:t+\omega-1}),
\end{equation}
where each future observation is conditioned on the observations at all previous timestamps.
In practice, the generative models may face some challenges when applied to real-world forecasting scenarios such as demand forecasting for online retailers.
In addition to the efficiency issue in both training and forecasting stages,
 there is also an error accumulation problem as each prediction is fed back as the ground-truth to forecast longer horizons, in which process errors may accumulate.
Instead of applying the classical generative approach, our framework forecasts the joint distribution of future observations directly:
\begin{equation}\label{direct}
    P\left({\mathbf y}_{(t+1):(t+\Omega)}|{\mathbf y}_{1:t}\right) =
    \prod\limits_{\omega=1}^{\Omega} p({\mathbf y}_{t+\omega}|{\mathbf y}_{1:t}).
\end{equation}
While time series data usually have systematic patterns such as trend and seasonality, it is also crucial that a forecasting framework allows covariates $X_{t+\omega}^{(i)}~(\mathrm{where}~\omega = 1,...,\Omega~\mathrm{and}~i = 1, ..., N)$ that include additional information to the direct forecasting strategy in Equation~\ref{direct}. The joint distribution of the future incorporating the covariates becomes:
\begin{equation}\label{direct-cov}
    P\left({\mathbf y}_{(t+1):(t+\Omega)}|{\mathbf y}_{1:t}\right) =
    \prod\limits_{\omega=1}^{\Omega} p({\mathbf y}_{t+\omega}|{\mathbf y}_{1:t}, X_{t+\omega}^{(i)}, i=1,...,N).
\end{equation}

Under the above settings, the challenge becomes to design a neural network framework that incorporates the historical observations ${\mathbf y}_{1:t}$ and  the covariates $X_{t+\omega}^{(i)}$. 
In the following sections, we describe how we extend the idea of the dynamic regression model (e.g., the ARIMAX model) to  build a direct forecasting framework for multiple time series by applying dilated causal convolutions and residual neural networks. We will then describe the probabilistic forecasting framework in detail and some practical considerations of the input features.

\subsection{Neural network architecture}
Dynamic regression models (e.g., ARIMAX) extend the classical time series model to include both information from past observations and exogenous variables (\citealt{pankratz2012forecasting}). A way to represent dynamic regression models is as follows:
\begin{equation*}
y_{t}^{(i)} = \nu_B(X_{t}^{(i)})+n_t^{(i)}.
\end{equation*}
where $\nu_B(\cdot)$ is a transfer function that describes how the changes in exogenous variables $X_{t}^{(i)}$ are transferred to $y_t^{(i)}$,
and $n_t^{(i)}$ is a stochastic time series process, e.g., the ARIMA process, which captures a forecast of $y_t^{(i)}$ using historical information.

To extend the dynamic regression model to multiple time series forecasting scenario, we propose a variant of residual neural network~\citep[resnet,][]{he2016deep,he2016identity}. Its main difference from the original resnet is that the new block allows for two inputs -- one input for the historical observations and the other for exogenous variables. 
for convenience, we refer it as resnet-v in the rest of the paper. Section~\ref{sec:decoder} provides more details of the module resnet-v.

%  in which the original resnet is described as:
% \begin{equation*}
%     \mathbf{z} = \mathcal{F}(\mathbf{x}, {W_i})+\mathbf{x}
% \end{equation*}

% The new block can be described as follows:
% \begin{equation*}
%     \mathbf{z} = \mathcal{R}(X_{t}^{(i)})+\mathbf{h_t^{(i)}}
% \end{equation*}
% And hence the residual function $R$ plays the role of transfer function in dynamic regression model
% and $h_t$ is the latent output from stochastic time series process.

\begin{figure*}[htp!]
  \centering
    \begin{subfigure}{0.9\textwidth}
        \includegraphics[height=7.cm]{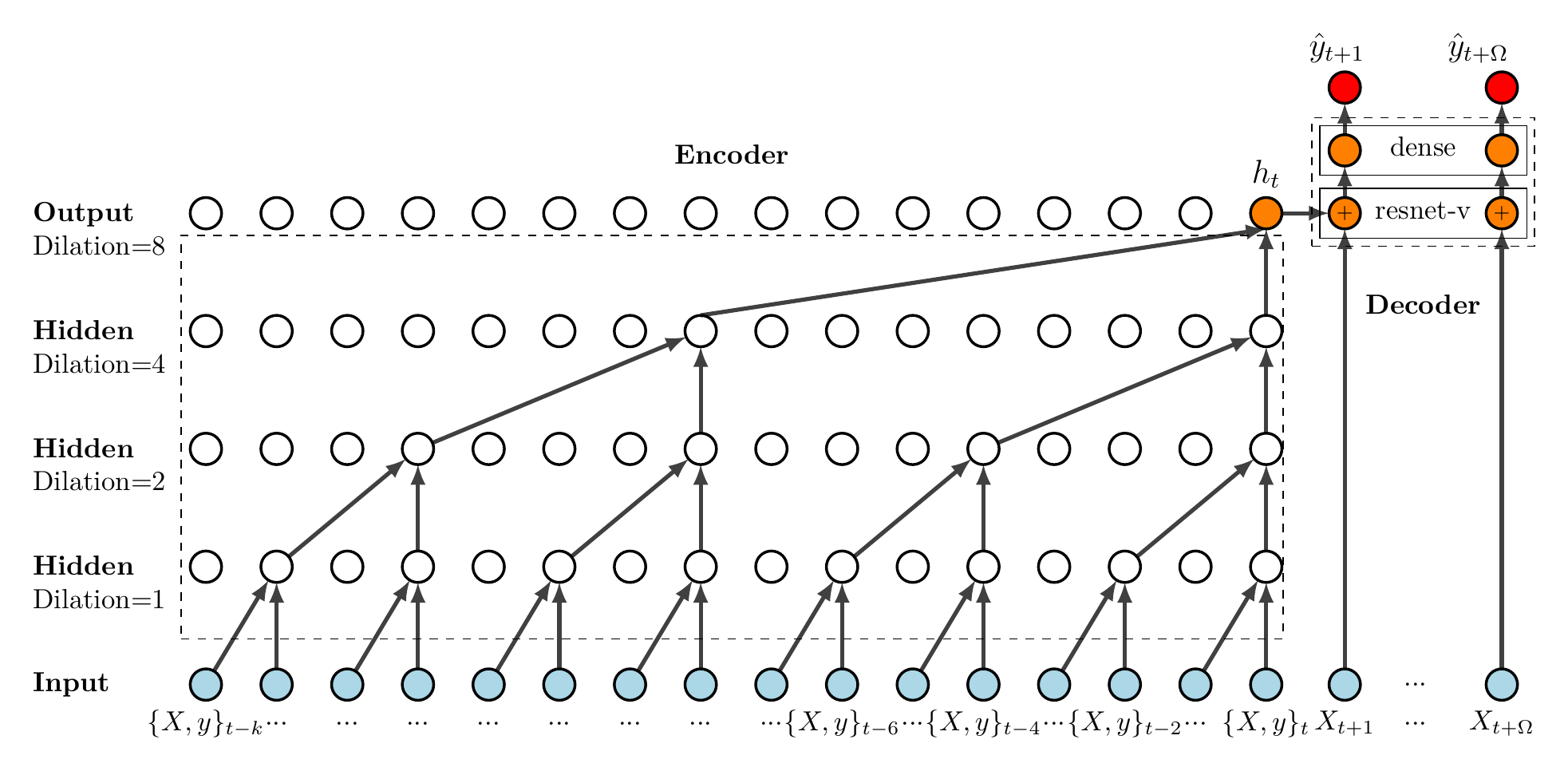}
        \caption{\small{Architecture of \textbf{DeepTCN}}}
        \label{fig:TCNFramework}
    \end{subfigure}
    \\
    \qquad \qquad
    \begin{subfigure}{0.4\textwidth}
  \includegraphics[height=7.cm]{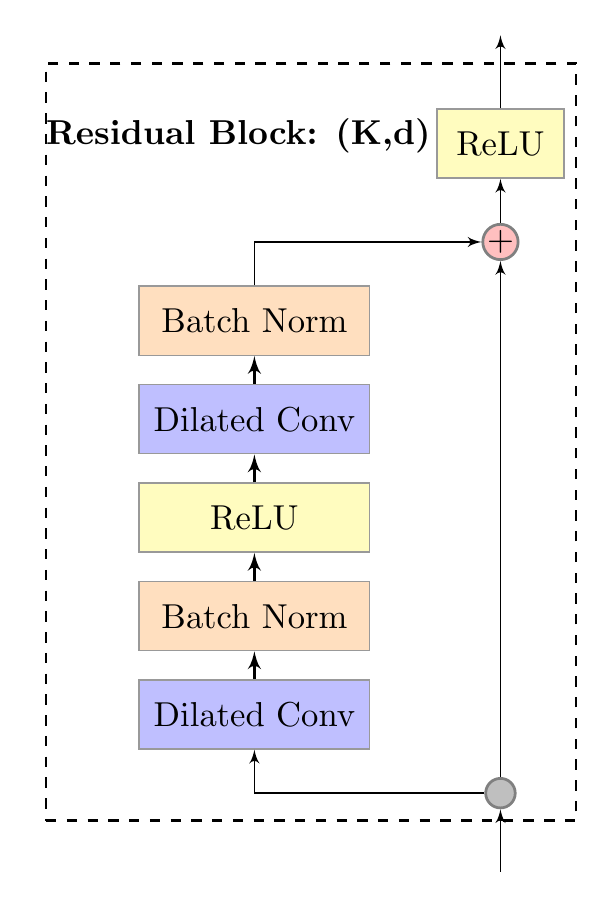}
  \caption{Encoder module}
      \label{fig:Encoder}
      \end{subfigure}
      \qquad \begin{subfigure}{0.4\textwidth}
  \includegraphics[height=7.cm]{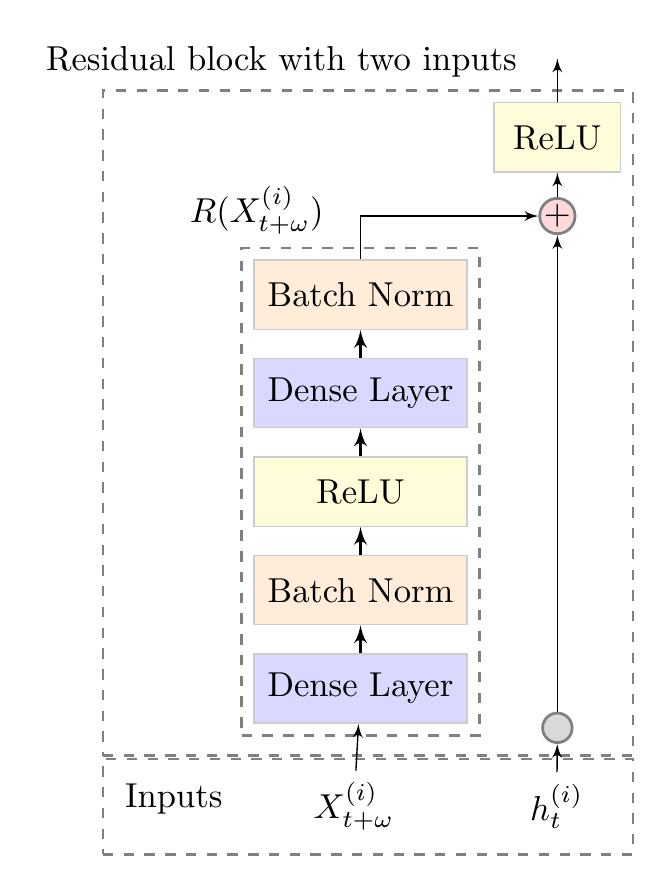}
    \caption{Decoder module}
      \label{fig:Decoder}
  \end{subfigure}
  \caption{(a) Architecture of DeepTCN. Encoder part: stacked dilated causal convolutional nets are constructed to capture the long-term temporal dependencies.
    Decoder part: the decoder includes a variant of residual block (referred as resnet-v, shown as $\oplus$) and an output dense layer. The module resnet-v is designed to integrate output of stochastic process of historical observations and future covariates. Then the output dense layer is adopted to map the output of resnet-v into our final forecasts.
  (b) Encoder module. Residual blocks are taken as the ingredient. Each residual block consists of two layers of dilated causal convolutions, the first of which is followed by a batch normalization and ReLU and the second of which is follow by another batch normalization. The output is taken as the input of  the residual block, followed by another ReLU. (c) Decoder module.  $h_{t}^{(i)}$ is the output of the encoder, $X_{t+\omega}^{(i)}$ are the future covariates, and $R(\cdot)$ is the nonlinear function applied on $X_{t+\omega}^{(i)}$. For the residual function $R(\cdot)$, we first apply a dense layer and a batch normalization to project the future covariates. Then a ReLU activation is applied followed by another dense layer and batch normalization.}
  %The decoder part produces the probabilistic estimation of interest.}
\end{figure*}

In this paper, we propose the Deep Temporal Convolutional Network (DeepTCN). The entire architecture of DeepTCN is presented in Figure~\ref{fig:TCNFramework}. The high-level architecture is similar to the classical Seq2Seq framework.
In the encoder part, stacked dilated causal convolutions are constructed to model the stochastic process of historical observations and output $h_t^{(i)}$.
Then, the module resnet-v in the decoder part incorporates the latent output  $h_t^{(i)}$ and future exogenous variables $X_{t+\omega}^{(i)}$, and outputs another latent output.
%Finally, a dense layer is applied to map the output of the variant of resnet and predicts the joint probability of future observations. 
Finally, a dense layer is applied to map the output of resnet-v and to produce the probabilistic forecasts of future observations. In the following sections, we provide further details for each module.

%Then the decoder part predicts the joint probability of future observations based on the decoder output $h_t^{(i)}$ and future exogenous variables $X_{t+\omega}^{(i)}$. 
%The decoder is designed in such a way for two reasons: 1) such a framework can naturally cooperate two parts of inputs: the outputs of encoder and the future covariates, and 2) from the perspective of time series modeling, a future observation can be considered to be composed of an autocorrelation component determined by past covariates and a nonlinear component determined by the future knowledge.
%In other words, the residuals between the future observations and predictions solely determined by the historical covariates can be explained as the function of future covariates.

%Here, our framework apply dilated causal convolutional to model the stochastic process.

\subsubsection{Encoder: Dilated causal convolutions}
Causal convolutions are convolutions where the output at time $t$ can only be obtained from the inputs that are no later than $t$.
Dilation causal convolutions allow the filter to be applied over an area larger than its length by skipping the input values with a certain step ~\citep{van2016wavenet}.
In the case of univariate series, given a single-dimensional input sequence $x$, the output (feature map) $s$ at location $t$ of a dilated convolution with kernel $w$ can be expressed as:
\begin{equation}
    s(t) = (x*_{d}w)(t) = \sum\limits_{k=0}^{K-1} w(k)x(t-d\cdot k),
\end{equation}
where $d$ is the dilation factor, and $K$ is the size of the kernel.
Stacking multiple dilated convolutions enable networks to have very large receptive fields and to capture long-range temporal dependencies with a smaller number of layers. The left part of Figure~\ref{fig:TCNFramework} is an example of dilated causal convolutions with dilation factors $d=\{1,2,4,8\}$, where the filter size $K=2$ and a receptive field of size $16$ is reached by staking four layers.

Figure~\ref{fig:Encoder} shows the basic module for each layer of the encoder, where both of two dilated convolutions inside the module have the same kernel size $K$ and dilation factor $d$.
Instead of implementing the classical gating mechanism in Wavenet~\citep{van2016wavenet}, in which a dilated convolution is followed by a gating activation, residual blocks are taken as the ingredient.
As shown in Figure~\ref{fig:Encoder}, each residual block consists of two layers of dilated causal convolutions, the first of which is followed by a batch normalization and rectified nonlinear unit (ReLU)~\citep{nair2010rectified} while the second of which is followed by another batch normalization~\citep{ioffe2015batch}. The output after the second batch normalization layer is taken as  the input of the residual block, followed by a second ReLU.
Residual blocks have been proven to help efficiently train and stabilize the network, especially when the input sequence is very long. More importantly, non-linearity gained by the rectified linear unit (ReLU) achieves better prediction accuracy in most of the empirical studies.
Various Natural Language Processing (NLP) tasks also support the above conclusion ~\citep{bai2018empirical}.

\subsubsection{Decoder: Residual neural network}
\label{sec:decoder}
The decoder includes two parts. 
The first part is the variant of residual neural network, the module resnet-v.  
The second part is a dense layer that maps the output of the resnet-v to the probabilistic forecasts.
As mentioned before, the module resnet-v
allows for two inputs (one for the historical information and the other for exogenous variables), and is designed to capture the information of these two inputs. It can be written as:
\begin{equation*}
    \delta_{t+\omega}^{(i)} = R(X_{t+\omega}^{(i)}) + {h_t^{(i)}},
\end{equation*}
where $h_t^{(i)}$ is the latent output by the encoder, $X_{t+\omega}^{(i)}$ are the future covariates and $\delta_{t+\omega}^{(i)}$ is the latent output of resnet-v. $R(\cdot)$ is the residual function applied on $X_{t+\omega}^{(i)}$.
Hence the nonlinear function $R(\cdot)$ plays the role of transfer function in dynamic regression model and explains the residuals between ground truth and predictions solely determined by the encoder part (e.g, the promotion effects on online retailer platforms or weather forecast for brick and mortar retailers).

Figure~\ref{fig:Decoder} shows the structure of the resnet-v.
For the residual function $R(\cdot)$, we first apply a dense layer and a batch normalization to project the future covariates. Then a ReLU activation is applied followed by another dense layer and batch normalization.
%Such a decoder also enjoys the flexibility to include additional features (e.g., promotion plans provided by operation teams or weather forecasts for brick and mortar retailers).
Finally, an output dense layer maps the latent variable $\delta_{t+\omega}^{(i)}$  to produce the final output $Z$ that corresponds to the probabilistic estimation of interest.

In the next section, we describe how we construct the probabilistic forecasting framework via neural networks in the output dense layer.

\subsection{Probabilistic forecasting framework}
Neural networks enjoy the flexibility to produce multiple outputs.
In the DeepTCN framework, for each future observation, the output dense layer in the decoder can produce $m$ outputs: $Z=(z^1,..., z^m)$, which represent the parameter set of the hypothetical distribution of interest.
Take Gaussian distribution as an example, for the $\omega$-th future observation of the $i$-th series, $y_{t+\omega}^{(i)}$, the output layer produces two outputs (the mean and the standard deviation), which gives $Z_{t+\omega}^{(i)}=(\mu_{t+\omega}^{(i)}, \sigma_{t+\omega}^{(i)})$, where $\mu_{t+\omega}^{(i)}$ is the expectation of $y_{t+\omega}^{(i)}$ and $\sigma_{t+\omega}^{(i)}$ is the standard deviation.
Therefore, the probabilistic forecasts can be described as:
\begin{equation}\label{cd1}
    P\left(y_{t+\omega}^{(i)} \right)\sim G(\mu_{t+\omega}^{(i)}, \sigma_{t+\omega}^{(i)}).
\end{equation}

More specifically, we consider two probabilistic forecasting frameworks in this paper. 
The first one is the parametric framework, in which probabilistic forecasts of future observations can be achieved by directly predicting the parameters of the hypothetical distribution (e.g., the mean and the standard deviation for Gaussian distribution) based on maximum likelihood estimation.
The second one is non-parametric, which produces a set of forecasts corresponding to quantile points of interest~\citep{koenker1978regression} with $Z$ representing the quantile forecasts.

In practice, whether to choose the parametric approach or the non-parametric approach depends on the application context.  The parametric approach requires the assumption of a specific probability distribution while the non-parametric approach is distribution-free and thus is usually more robust. However, a decision-making scenario may rely on the sum of probabilistic forecasts for a certain period. For example, an inventory replenishment decision may depend on the distribution of the sum of demand for the next few days. In such cases, the non-parametric approach will not work since the output (e.g., the quantiles) is not additive over time and the parametric approach has its advantage of being flexible in obtaining such information by sampling from the estimated distributions.

\subsubsection{Non-parametric approach}
In the non-parametric framework, forecasts can be obtained by quantile regression. In quantile regression~\citep{koenker1978regression}, denoting the observation and the prediction for a specific quantile level $q$ as  $y$ and $\hat y^{q}$  respectively, models are trained to minimized the quantile loss, which is defined as
\begin{equation}
    L_q(y, \hat y^{q} ) = q(y - \hat y^{q})^{+}+(1-q)(\hat y^{q} - y)^{+},
    \label{quantileloss}
\end{equation}
where $(y)^{+} = \max(0,y)$ and $q \in (0,1)$. Given a set of  quantile levels  $Q=(q_1,...,q_m)$, the $m$ corresponding forecasts can be obtained by minimizing the total quantile loss defined as
\begin{equation*}
    L_{Q} =  \sum\limits_{j=1}^m L_{q_j} \left(y, \hat y^{q_j} \right).
\end{equation*}

\subsubsection{Parametric approach}
For the parametric approach, given the predetermined distribution (e.g., Gaussian distribution), the maximum likelihood estimation is applied to estimate the corresponding parameters.
Take Gaussian distribution as an example, for each target value $y$, the network outputs the parameters of the distribution, namely the mean and the standard deviation, denoted by $\mu$ and $\sigma$, respectively.
The negative log-likelihood function is then constructed as the loss function:
\beqa
 L_{G} &=& -\log\ell (\mu,\sigma\,| y) \n
& = & -\log \left( (2\pi \sigma^2)^{-1/2}\exp\left[-(y-\mu)^2/(2\sigma^2)  \right] \right)\n
&=&\frac{1}{2}\log(2 \pi )+\log(\sigma) + \frac{(y-\mu)^2}{2\sigma^2}. \nonumber
\eeqa
We can extend this approach to a variety of probability distribution families. For example, we can choose negative-binomial distribution for long-tail products, which is traditionally used for modeling over-dispersed count data  and has been shown to perform well in empirical studies \citep{VILLANI2012121, SNYDER2012485, SYNTETOS20151746, flunkert2017deepar}.

It is worth mentioning that some parameters of a certain distribution (e.g.,  the standard deviation in Gaussian distribution) must satisfy the condition of positivity. To accomplish this, we apply ``Soft ReLU'' activation, namely the transformation $ \hat z = \log (1+\exp(z))$, to ensure positivity~\citep{flunkert2017deepar}.

\subsection{Input features}
% two type of covariates
There are typically two kinds of input features: time-dependent features (e.g., product price and day-of-the-week) and time-independent features (e.g., product\_id, product brand and category).
% series specific covariates
Time-independent covariates such as product\_id contain series-specific information. 
Including these covariates helps capture the scale level and seasonality for each specific series.

To capture seasonality, we use hour-of-the-day, day-of-the-week, day-of-the-month for hourly data, day-of-the-year for daily data and month-of-year for monthly data. 
Besides, we use hand-crafted holiday indicators for shopping festival such as ``11.11'', which enables the model to learn spikes due to scheduled events.

Dummy variables such as product\_id and day-of-the-week are mapped to dense numeric vectors via embedding~\citep{mikolov2013distributed, mikolov2013efficient}. We find that the model is able to learn more similar patterns across series by representation learning and thus improves the forecasting accuracy for related time series, which is especially useful for series with little or no historical data.
In the case of new products or new warehouses without sufficient historical data, we perform zero padding to ensure the desired length of the input sequence.  

\section{Experiments}\label{experiment}
\subsection{Datasets}

\begin{table}[!thb]
\footnotesize
    \begin{center}
        %\begin{tabular}{llllll}
        \begin{tabular}{lccccc}
            \toprule
            & \texttt{JD-demand} & \texttt{JD-shipment}   & \texttt{electricity} & \texttt{traffic} & \texttt{parts} \\ \midrule
        Number  & 50,000 & 1,450 & 370  & 963   & 1,406\\ %\hline
        Length  & $[0,1800]$  & $[0,1800]$ & 26,304 & 10,560 & 51 \\
        Domain  & $\N$ &  $\N$ & $\R^{+}$ &  $[0,1]$ & $\N$\\ 
        Granularity &daily & daily & hourly & hourly &  monthly  \\ \hline
    \end{tabular}
    \end{center}
    \caption{Summary of the datasets used in the experiments.}
    \label{table:dataset}
\end{table}

We evaluate the performance of DeepTCN on five datasets. 
More specifically, within the DeepTCN framework, two models -- the non-parametric model that predicts the quantiles and the parametric Gaussian likelihood model -- are applied for the forecasting performance evaluation. We refer to them as \texttt{DeepTCN-Quantile} and \texttt{DeepTCN-Gaussian}, respectively, for the rest of the paper.

Table~\ref{table:dataset} shows the details of the five datasets. \texttt{JD-demand} and \texttt{JD-shipment} are from JD.com, which correspond to two forecasting tasks for online retailers, namely demand forecasting of regional product sales and shipment forecasting of the daily delivery volume of packages in retailers' warehouses. 
Since it is inevitable for new products or warehouses to emerge, the training periods for these two datasets can range from zero to several years and the corresponding forecasting tasks involve situations such as cold-starts and data sparsity.  We also use three public datasets that have been widely used in various time series forecasting studies for accuracy comparison, namely 
 \texttt{electricity},  \texttt{traffic} and \texttt{parts}.
 The \texttt{electricity} dataset contains hourly time series of the electricity consumption of 370 customers.
 The \texttt{traffic} dataset is a collection of the occupancy rates (between 0 and 1) of 963 car lanes from San Francisco bay area freeways.
 The \texttt{parts} dataset is comprised of 1,046 time series representing monthly demand of spare parts in a US car company.
 A more detailed description of these datasets can be found in Appendix \ref{appendix:dataset}.

\subsection{Accuracy comparison}

Current baseline models for JD.com's datasets include seasonal ARIMA (\texttt{SARIMA}) and \texttt{lightGBM}, a gradient boosting tree method that has been empirically proven to be a highly effective approach in predictive modeling. These two online models are deployed and continuously improved to provide more accurate forecasts and to better serve the consequent business operations (e.g., inventory replenishment).
A more detailed description including the features and parameters used in these two models can be found in Appendix \ref{appendix:baseline}. For the public datasets, we compare \texttt{DeepTCN} with \texttt{DeepAR}~\citep{flunkert2017deepar} as implemented using the student-$t$ distribution~\citep{gasthaus2019probabilistic}, \texttt{SQF-RNN}~\citep{gasthaus2019probabilistic} and \texttt{DeepState}~\citep{rangapuram2018deep}.

\subsubsection{Evaluation metrics}
\label{appendix:metrics}

To evaluate probabilistic forecasting, given $N$ time series $\{y^{(i)}\}_{i=1}^N$ and the prediction range $\{t+1, t+2,...,t+\Omega\}$, we use the normalized sum of quantile losses \citep{gasthaus2019probabilistic}, which is denoted as 
\begin{eqnarray*}
QL_q =  \frac{\sum_{i,t} L_q(y_t^{(i)}, \hat{y}_t^{(i)})}{\sum_{i,t}|y_{t}^{(i)}|},
\end{eqnarray*}
where $L_q(\cdot)$ is defined in Equation~\ref{quantileloss}. We refer to $QL_q$ as the $q$-quantile loss.
% \begin{eqnarray*}
%   P_q(y,\hat y) = \left\{ \,
%     \begin{IEEEeqnarraybox}[][c]{l?s}
%       \IEEEstrut
%         q(y-\hat y) & if $y>\hat y$, \\
%         (1-q)(\hat y -y ) & otherwise.
%       \IEEEstrut
%     \end{IEEEeqnarraybox}
%     \right.
%   \label{eq:example_left_right1}
% \end{eqnarray*}

The evaluation metrics used in our experiments for point forecasting include the Symmetric Mean Absolute Percent Error (SMAPE), the Normalized Root Mean Square Error (NRMSE) and the Mean Absolute Scaled Error (MASE), which are defined as follows. Note that for the MASE, the value $m$ is the seasonal frequency.
\beqa
SMAPE &=& \frac{1}{N(\Omega-t)}\sum\limits_{i, t} \left\vert \frac{2\left(y_{t}^{(i)}-\hat y_{t}^{(i)}\right)}{y_{t}^{(i)}+\hat y_{t}^{(i)}}\right\vert, \n
% RMSLE &=& \sqrt{\frac{1}{N(\Omega-t)}\sum_{i, t} \left(\log\left(y_{t}^{(i)}+1\right)-\log\left(\hat y_{t}^{(i)}+1\right)\right)^2}, \n
%  ND & =& \frac{\sum_{i,t} \left|y_{t}^{(i)}-\hat y_{t}^{(i)}\right|}{\sum_{i,t} \left|y_{t}^{(i)}\right|}, \n
NRMSE &=& \frac{\sqrt{\frac{1}{N(\Omega-t)}\sum\limits_{i,t} \left(y_{t}^{(i)}-\hat y_{t}^{(i)}\right)^2}}{\frac{1}{N(\Omega-t)}\sum\limits_{i,t} \left|y_{t}^{(i)}\right|}, \nonumber \n
MASE &=& \frac{1}{N(\Omega-t)}\sum\limits_{i}\frac{\sum\limits_{t} \left|y_{t}^{(i)}-\hat y_{t}^{(i)}\right|} {\frac{1}{T-m}\sum\limits_{t=m+1}^T \left|y_{t}^{(i)}- y_{t-m}^{(i)}\right|}, \nonumber
\eeqa
where $y_{t}^{(i)}$ is the true value of series $i$ at time step $t$, and $\hat y_{t}^{(i)}$ is the corresponding prediction value.
    %and $N$ is the number of all points in the testing periods.

\subsubsection{Results on JD.com's datasets}

We start with comparing the probabilistic forecasting results of \texttt{DeepTCN} against the online \texttt{SARIMA} and \texttt{lightGBM} models on JD.com datasets over two testing periods: Oct 2018 and Nov 2018. In particular, China's largest shopping festival ``11.11'' lasts from Nov 1 to Nov 12, during which Nov 11 is the biggest promotion day. We use the 0.5-quantile loss and 0.9-quantile loss as the evaluation metrics, which are referred to as QL50 and QL90, respectively.
The model \texttt{DeepTCN-Quantile} is trained to predict $q$-quantiles with $q \in \{0.5, 0.9\}$.
For the model \texttt{DeepTCN-Gaussian}, the quantile predictions are obtained by calculating the percent point function of Gaussian distribution (the inverse of cumulative density function) at $0.5$ and $0.9$ quantiles.

The comparison results of \texttt{JD-demand} and \texttt{JD-shipment} are illustrated in Table \ref{Table:jdcom0}.
As we can see, both \texttt{DeepTCN-Quantile} and \texttt{DeepTCN-Gaussian} achieve better results than the two online models.  In particular, \texttt{DeepTCN-Quantile} performs the best. 
One possible reason is that \texttt{DeepTCN-Gaussian} is constructed based on the Gaussian likelihood, but these datasets do not necessarily follow the assumption of normal distribution.
On the contrary, the model \texttt{DeepTCN-Quantile}, in light of the distribution-free nature,  generates better forecasts by minimizing the quantile loss directly.

\begin{table}
 \begin{center}
    \linespread{1.3}\footnotesize
    \begin{tabular}{lccccc}
    \toprule
     Method &\multicolumn{2}{c} {\texttt{JD-demand}} &&\multicolumn{2}{c} {\texttt{JD-shipment}} \\
     \cline{2-3}\cline{5-6}
             & Oct 2018 & Nov 2018 && Oct 2018 & Nov 2018 \\
             \midrule
            %\texttt{SARIMA} &  0.753/0.626 & 0.771/0.965 && 0.307/0.197 & 0.398/0.315 \\
            \texttt{SARIMA} &  0.377/0.313 & 0.385/0.482 && 0.154/0.098 & 0.199/0.158 \\
            %\texttt{lightGBM} &  0.731/0.607 & 0.784/0.988 && 0.2880/0.173 & 0.411/0.271 \\
            \texttt{lightGBM} &  0.366/0.303 & 0.392/0.494 && 0.144/0.087 & 0.205/0.136 \\
            %\texttt{DeepTCN-Quantile} & \bf{0.653/0.528} & \bf{0.698/0.701} && \bf{0.173/0.100} & \bf{0.247/0.160} \\
            \texttt{DeepTCN-Quantile} & \bf{0.328/0.264} & \bf{0.349/0.350} && \bf{0.087/0.050} & \bf{0.124/0.080} \\
            %\texttt{DeepTCN-Gaussian} & 0.697/0.588 & 0.720/0.873 && 0.188/0.105 & 0.326/0.219 \\        
            \texttt{DeepTCN-Gaussian} & 0.349/0.294 & 0.360/0.436 && 0.094/0.052 & 0.163/0.109 \\        
     \bottomrule
    \end{tabular}
    \end{center}
    \caption{Comparison of probabilistic forecasts on \texttt{JD-demand} and \texttt{JD-shipment} datasets. The quantile losses QL50/QL90 are evaluated against online models over two testing periods -- Oct 2018 and Nov 2018.}
    \label{Table:jdcom0}
\end{table}

We then present in Table~\ref{table:com0} an accuracy comparison of point forecasting between our model and the other two baseline models including \texttt{SARIMA} and \texttt{lightGBM}.
The point forecasting results of \texttt{DeepTCN-Quantile} are achieved with the non-parametric approach that predicts  the $0.5$ quantiles. 
In Table~\ref{table:com0}, \texttt{All-Data} consists of all series in the dataset; \texttt{Long-series} includes series with historical data longer than two years; \texttt{Short-series} are those starting after 2018, i.e., without historical shopping festival data. We can see that \texttt{DeepTCN-Quantile} achieves consistently the best accuracy with regard to all the metrics across all data groups. In particular, when historical shopping festival data is not available, the performance of \texttt{SARIMA} and \texttt{lightGBM} become much worse (the result in \texttt{Short-series}), while \texttt{DeepTCN-Quantile} maintains the same performance level.

\begin{table}
    \begin{center}
    \begin{tabular}{llccc} 
     \toprule
     Data group   & Method & NRMSE& SMAPE & MASE  \\ 
      \midrule
     \multirow{3}{*}{\texttt{All-data}} &   \texttt{SARIMA}&1.285  & 0.369 & 0.911 \\ 
                                        & \texttt{lightGBM} &1.244 & 0.430 & 0.918 \\
                                        & \texttt{DeepTCN-Quantile} & \textbf{0.873} & \textbf{0.284} & \textbf{0.763} \\
             \midrule 
     \multirow{3}{*}{\texttt{Long-series}} &   \texttt{SARIMA} &1.044 & 0.323 & 0.900 \\ 
                                           & \texttt{lightGBM} & 0.995& 0.312 & 0.903 \\
                                           & \texttt{DeepTCN-Quantile}  &\textbf{0.895} & \textbf{0.268} & \textbf{0.708} \\
             \midrule 
     \multirow{3}{*}{\texttt{Short-series}} &   \texttt{SARIMA} &1.376 &  0.430 & 0.980 \\ 
                                            & \texttt{lightGBM} &1.536 & 0.457 & 1.096 \\
                                            & \texttt{DeepTCN-Quantile}  &\textbf{0.923} & \textbf{0.354} & \textbf{0.798} \\
             \bottomrule
               \end{tabular}
    \end{center}
    \caption{Point forecasting accuracy comparison on NRMSE, SMAPE and MASE of different subgroups of \texttt{JD-shipment} in Nov 2018. \texttt{All-Data} represents all series with the length of training periods ranging from zero to four years; \texttt{Long-series} includes the warehouses with historical data of more than two years; \texttt{Short-series} indicates series starting after 2018, namely those with no historical shopping festival data.}
    \label{table:com0}
\end{table}

\begin{figure*}
  \centering
    \includegraphics[width=1\textwidth]{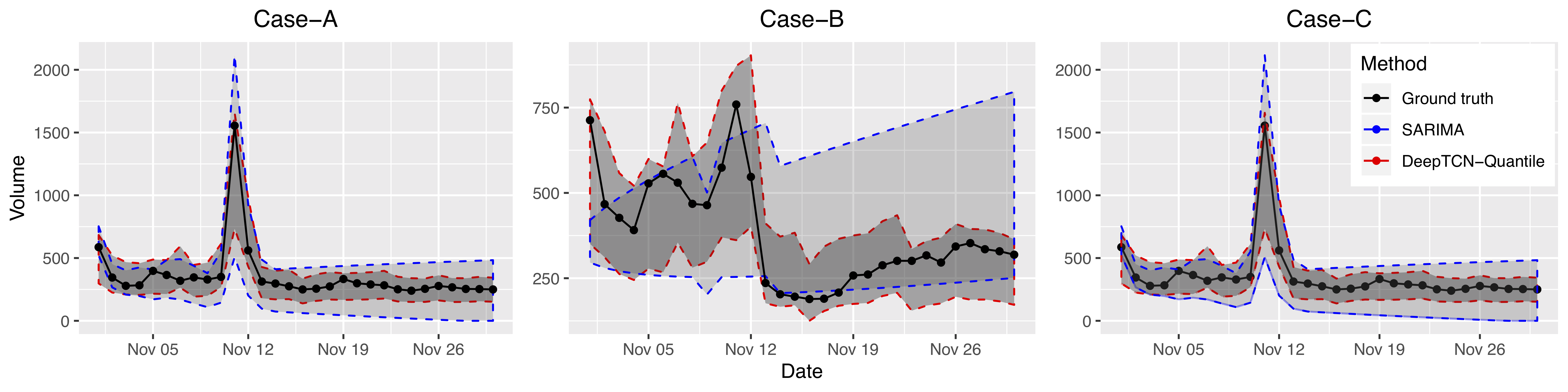}
    \caption{Probabilistic forecasts of \texttt{SARIMA} and \texttt{DeepTCN-Quantile} for three cases (randomly chosen for illustration purposes). Case A and Case B show the forecasting results of two fast-moving products; Case C shows the forecasting results of the daily delivery volume of packages from one warehouse. The ground truth, and the [10\%, 90\%] prediction intervals of \texttt{SARIMA} and \texttt{DeepTCN-Quantile} are also shown in different colors. (For interpretation of the references to colour in this figure, the reader is referred to the web version of this article.)}
    \label{figure:case}
\end{figure*}

To gain a better understanding of the performance improvement exhibited by the proposed DeepTCN framework, we show in Figure~\ref{figure:case} three cases of probabilistic forecasts generated by \texttt{SARIMA} and \texttt{DeepTCN-Quantile}. Case A and Case B are two demand forecasting examples of Oct 2018 and Nov 2018, respectively,  while Case C is an example of shipment forecasting of Nov 2018. It is shown that for both tasks, \texttt{DeepTCN-Quantile} generates more accurate uncertainty estimation. 
Moreover, \texttt{SARIMA} postulates increasing uncertainty over time while the uncertainty estimation of \texttt{DeepTCN-Quantile} is well learned from the data. For example, the uncertainty of \texttt{SARIMA} during the shopping festival period is huge due to both promotion activities and intense market competition.

\begin{figure*}[thb!]
  \centering
  \includegraphics[width=1\textwidth]{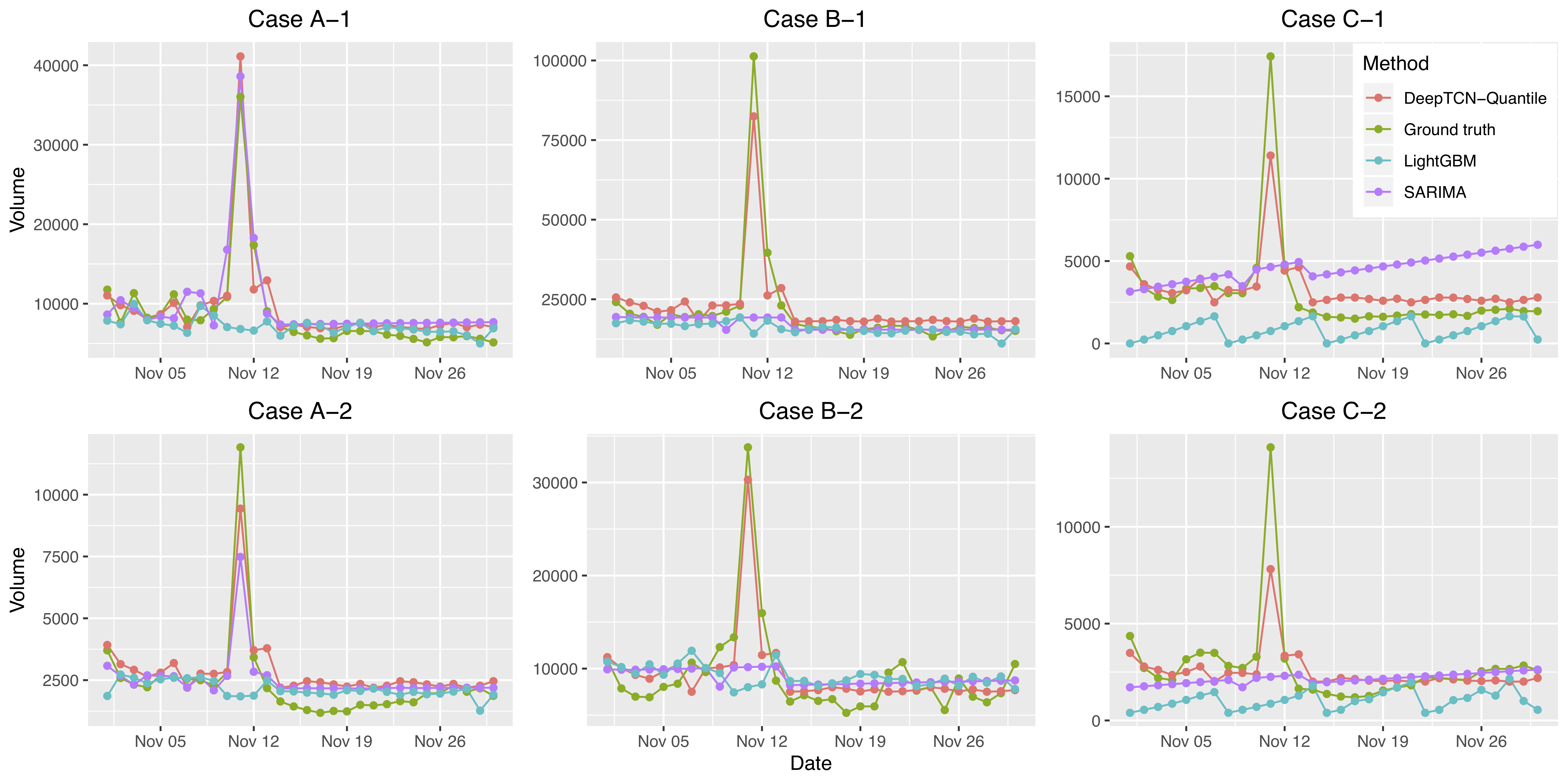}
    \caption{Point forecasts of \texttt{DeepTCN-Quantile}, \texttt{SARIMA} and \texttt{lightGBM} for six cases (randomly chosen from \texttt{JD-shipment} for illustration purposes). Cases A-1 and A-2 are examples with historical data of more than two years;
    cases B-1 and B-2  show instances without previous shopping festival data;  cases C-1 and C-2 illustrate cold-start forecasting namely the forecasting of time series with little historical data,  e.g., less than three days. 
    Note that Nov 11 is one of China's biggest promotion days. (For interpretation of the references to colour in this figure, the reader is referred to the web version of this article.)}
    \label{figure:case2}
\end{figure*}

Finally, we perform a qualitative analysis on \texttt{JD-shipment} dataset over the testing period of Nov 2018. 
We choose this dataset because 1) it consists of series whose magnitudes of volume are high and stable, and  2) the testing period involves China's biggest shopping festival ``11.11''. As mentioned before, the occurrence of this festival results in a spike for the shipment volume, in which the forecasting tasks become more challenging. In Figure~\ref{figure:case2}, we illustrate cases of point forecasting under three different scenarios.
``11.11'' is the major promotion day and we can observe a spike in the true volume. In Cases A-1 and A-2 , where historical data of more than two years is available, all models can learn a similar volume pattern, including the spike on ``11.11''. However, \texttt{SARIMA} and \texttt{lightGBM} in Cases B-1 and B-2 fail to capture the spike on ``11.11'' due to lack of sufficient training data for historical festivals.  
Cases C-1 and C-2 are selected to demonstrate how these models handle cold-start forecasting. It turns out that \texttt{DeepTCN-Quantile}  stands out for this situation as it is able to capture both scale and shape patterns of the new warehouses by learning data from other warehouses with similar store-specific features.

\subsubsection{Results on the public datasets}

%We compare \texttt{DeepTCN} against \texttt{DeepAR}~\citep{flunkert2017deepar},\texttt{SQF-RNN}~\citep{gasthaus2019probabilistic} and \texttt{DeepState}~\citep{rangapuram2018deep}, which got the strongest published results on these datasets. 
%We also report the results of classical forecasting methods including \texttt{auto}.\texttt{arima} and \texttt{ETS}. Both methods are implemented in \textbf{R}'s \texttt{forecast} package~\citep{JSSv027i03}.

    %We start with conducting the experiments of probabilistic forecasting.    
Our first experiment on public datasets is to evaluate the performance of DeepTCN regarding probabilistic forecasting.
For \texttt{electricity} and \texttt{traffic} dataset, we implement a 24-hour ahead forecasting task for last seven days based on a rolling-window approach as described in~\citet{flunkert2017deepar}. 
For \texttt{parts} dataset, we evaluate the performance for last 12 months.
It is worth noting that we use the same model trained on the data before the first prediction window rather than retraining the model after updating the forecasts. 
In all forecasting experiments, we train the \texttt{DeepTCN-Quantile} models to predict $q$-quantiles with $q \in \{0.5, 0.9\}$,
and for the model \texttt{DeepTCN-Gaussian}, the quantile predictions are obtained by calculating the percent point function of Gaussian distribution at 0.5 and 0.9 quantiles, which is the same approach as applied in JD.com's datasets.

    %We compare our models against \texttt{DeepAR}~\citep{flunkert2017deepar},\texttt{SQF-RNN}~\citep{gasthaus2019probabilistic} and \texttt{DeepState}~\citep{rangapuram2018deep}, which got the strongest published results on these datasets.
%We also report the results of classical forecasting methods including \texttt{auto}.\texttt{arima} and \texttt{ETS}. Both methods are implemented in \textbf{R}'s \texttt{forecast} package~\citep{JSSv027i03}.

\begin{table*}[!thb]
    \centering
    \begin{center}\footnotesize
    \setlength\tabcolsep{4pt}
    \resizebox{\linewidth}{!}{
    \begin{tabular}{l ccccccccc}
     \toprule
     Dataset & \texttt{ETS} & \texttt{auto.arima} & \texttt{DeepAR-$t$} & \texttt{SQF-RNN}&  \texttt{DeepState} & \texttt{DeepTCN-Quantile} & \texttt{DeepTCN-Gaussian}  \\ \midrule 
     \texttt{electricity} &0.100/0.050 & 0.142/0.054& 0.068/0.033 &0.066/0.035 &\bfseries 0.043/0.025 & 0.057/0.029 & 0.062/0.039\\
     %\texttt{electricity} &0.121/0.101 & 0.283/0.109& 0.153/0.147 & &\bfseries 0.087/0.050 & 0.114/0.058 & 0.124/0.078\\
 %    \texttt{traffic} & 0.621/0.650 &0.492/0.280 & 0.177/0.153 & & 0.168/0.117 &  \bfseries{0.115/0.079}& \bfseries{0.141/0.097} \\
     \texttt{traffic} & 0.360/0.325 &0.246/0.140 & 0.117/0.090 & 0.119/0.090& 0.084/0.058 &  \bfseries{0.058/0.040}& \bfseries{0.071/0.049} \\
     %\texttt{parts} & 1.639/1.009& 1.644/1.066& 1.273/1.086 &--/-- &1.470/0.935 &  \bfseries{1.066/0.923}& \bfseries{1.245/0.930} \\
     \texttt{parts} & 0.820/0.505& 0.822/0.533 & 0.636/0.543 &---/--- &0.735/0.467 &  \bfseries{0.533/0.462}& \bfseries{0.623/0.465} \\
        \bottomrule
            \end{tabular}
            }
    \end{center}
    \caption{Accuracy comparison of probabilistic forecasting on public datasets. The numbers in the table are the QL50/QL90 results.}
    \label{table:publicCom0}
\end{table*}

Table \ref{table:publicCom0} illustrates the probabilistic forecasting results obtained by these models. It can be seen that the probabilistic forecasting results of both \texttt{DeepTCN-Quantile} and \texttt{DeepTCN-Gaussian} outperform other state-of-the-art models on \texttt{traffic} and \texttt{parts} datasets.
One possible reason is that there exists high correlation within these two datasets, and \text{DeepTCN} takes an advantage by learning the non-linear correlation among series, while traditional models such as \texttt{ETS} and \texttt{auto.arima} could not learn the shared patterns across the time series.

%For \texttt{electricity} dataset containing series that are not so related, \texttt{DeepState} achieves the best results and the performance of \texttt{DeepTCN} is slightly worse. We believe that models such as \texttt{DeepState} and \texttt{ES-RNN}~\citep{makridakis2018m4} have more advantages on situations where time series are not highly correlated as they specify unique parameters for each series.

\begin{table}[htp!]
    \centering
    \begin{center}
 \begin{tabular}{llcccc} \toprule
     Dataset & Method & NRMSE & SMAPE & MASE \\ \midrule 
     \texttt{electricity} & \texttt{ETS}  & 0.838 &0.156 & 1.247 \\
                          & \texttt{DeepAR-$t$}  & 0.550 & \textbf{0.110} &0.931 \\
      & \texttt{SQF-RNN}  & \textbf{0.518} &0.113 &0.937 \\
      & \texttt{DeepTCN-Quantile}  & 0.523 & 0.118 &  \textbf{0.926}\\ \midrule
      \texttt{traffic} & \texttt{ETS}  & 0.872  &0.594& 1.881 \\
                       & \texttt{DeepAR-$t$}  & 0.396 & \textbf{0.104} & 0.442 \\
      & \texttt{SQF-RNN}  & 0.381 & 0.117 & 0.449 \\
      & \texttt{DeepTCN-Quantile}  & \textbf{0.364} & 0.110 & \textbf{0.438} \\
     \bottomrule
 \end{tabular}
    \end{center}
    \caption{Accuracy comparison of point forecasting on public datasets.}
    \label{publicCom1}
\end{table}
To evaluate the performance of point forecasting, we comparing \texttt{DeepTCN} (\texttt{DeepTCN} that predicts the 0.5 quantiles) against \texttt{ETS}, 
%\texttt{DeepAR}~\citep{flunkert2017deepar} which implemented using the studen\Omega-t distribution~\citep{gasthaus2019probabilistic} and \texttt{SQF-RNN}~\citep{gasthaus2019probabilistic}.
\texttt{DeepAR-$t$}~\citep{flunkert2017deepar} and \texttt{SQF-RNN}~\citep{gasthaus2019probabilistic} 
and calculate the metrics NRMSE, SMAPE and MASE over the \texttt{electricity} and \texttt{traffic} datasets.
%In particular, we calculate the Normalized Root Mean Square Error (NRMSE), the Symmetric Mean Absolute Percentage Error (SMAPE) and the Mean Absolute Scaled Error (MASE) over the \texttt{electricity} and \texttt{traffic} datasets.
%In particular, we calculate the Normalized Root Mean Square Error (NRMSE), the Symmetric Mean Absolute Percentage Error (SMAPE) and the Mean Absolute Scaled Error (MASE) over the \texttt{electricity} and \texttt{traffic} datasets.
As can be seen from Table~\ref{publicCom1}, for \texttt{traffic} dataset with highly correlated series,
\texttt{DeepTCN-Quantile} achieves the best forecasting accuracy on NRMSE and MASE, while it performs the best on MASE for \texttt{electricity}.

\begin{figure*}[htp!]
  \centering
    \includegraphics[width=1\textwidth]{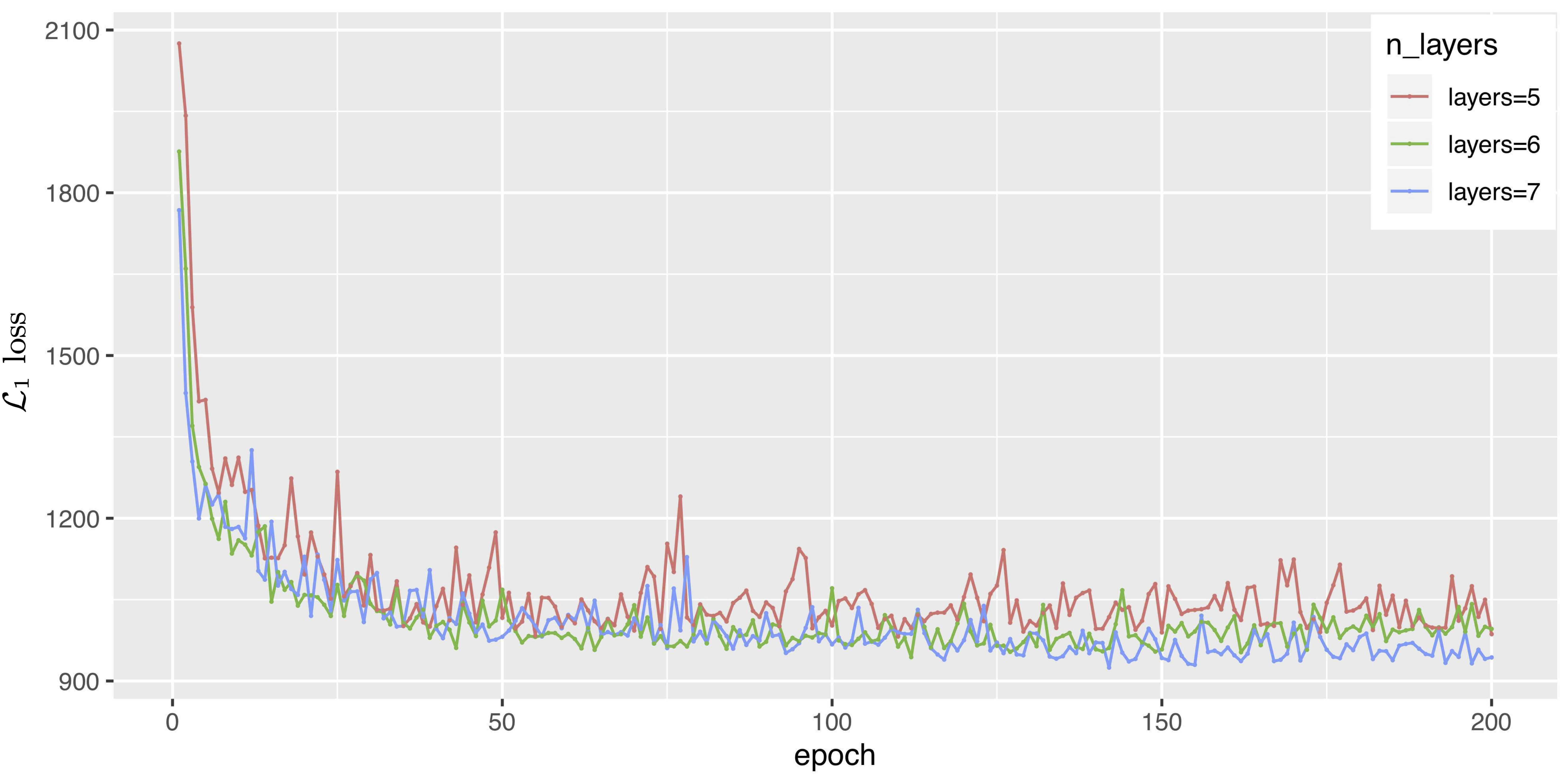}
    \caption{
        $\mathcal{L}_1$ loss over 200 epochs of three different architectures with 5, 6 and 7 encoder layers.}
    \label{figure:sensitivity}
\end{figure*}

\subsection{Sensitivity analysis}
Given that the stochastic process of historical observations is modeled by the stacked dilated causal convolutions in the encoder part of our DeepTCN framework, we now perform sensitivity analysis taking the \texttt{traffic} dataset as an example, to explore the effect of the number of encoder layers on the model performance.
In this experiment, we set the filter size of the dilated causal convolutions as $k=2$ and implement three model architectures: (1)
a 5-layer architecture with dilation factors $d
 = \{1,2,4,8,16\}$, (2) a 6-layer architecture with dilation factors $d = \{1,2,4,8,16,32\}$ and (3) a 7-layer architecture with dilation factors $d = \{1,2,4,8,16,20,32\}$.
Notice that in our experiment of traffic dataset, the length of input sequence is  $7 \times 24=168$, which is the hourly data of the previous week.
And for each layer of the dilated causal convolutions,
the kernel size times the dilation factor cannot exceed the length of the input sequence, thus we set the dilation factor of the third model as $d = \{1,2,4,8,16,20,32\}$ to ensure the input length of each layer is sufficient.

Figure~\ref{figure:sensitivity} shows the $\mathcal{L}_1$ loss of these three models over 200 epochs. It can be seen that  both the 6-layer and 7-layer architectures perform  better than the 5-layer architecture.
One reason is that the 5-layer architecture is relatively shallow to fully model the information from the historical observations. However, as one can see, the difference between using 6-layer and 7-layers is small, meaning that as long as one uses enough number of layers, the difference in result is quite small. Such phenomenon is quite consistent across other test cases. Thus our method is quite robust with respect to the model parameters.

%The  6-layer architecture performs similarly with 7-layer architecture in the first 150 epochs. However, the $\mathcal{L}_1$ loss of the 7-layer architecture continue  decreasing after 150 epochs. We believe that the model can learn more information from historical series and gain better forecasting accuracy with deeper layers.

%    Finally, we demonstrate in Table \ref{table:running time} a comparison with respect to run-time efficiency between  \texttt{DeepTCN} and \texttt{DeepAR}. Running times are obtained from the measurement of an end-to-end evaluation on datasets \texttt{electricity}, \texttt{traffic} and \texttt{parts},  including processing features, training the model, and producing the corresponding results. For \texttt{DeepTCN}, we show the run-time result of \texttt{DeepTCN-Quantile}.  For  \texttt{DeepAR}, we report the running time presented in \citet{flunkert2017deepar}. Both models are trained on the same GPU service \texttt{Tesla P40}.  
%As shown in Table \ref{table:running time}, \texttt{DeepTCN}, due to its capability of performing the convolutions in parallel, has a clear advantage on the run-time efficiency.

%\begin{table}[!hb]
%    \centering
%    \begin{center}
%    \begin{tabular}{lcc}
%     \toprule
%         Dataset & \texttt{DeepTCN} & \texttt{DeepAR}  \\ \midrule
%        \texttt{electricity} &  50m & 7h \\
%          \texttt{traffic} &  30m & 3h \\
%          \texttt{parts} & 40s & 5m \\\bottomrule 
%    \end{tabular}
%    \end{center}
%    \caption{Computation time comparison on public datasets.}
%    \label{table:running time}
%\end{table}

\section{Conclusion}\label{conclusion}
We present a convolutional-based probabilistic forecasting framework for multiple related time series and show both non-parametric and parametric approaches to model the probabilistic distribution based on neural networks. 
Our solution can help in the design of practical large-scale forecasting applications, which involves situations such as cold-starts and data sparsity. 
Results from both industrial datasets and public datasets show that the framework yields superior performance compared to other state-of-the-art methods in both point and probabilistic forecasting.

\section*{Acknowledgements}

Yanfei Kang's research were supported by the National Natural Science Foundation of China (No.~11701022). 
\section*{References}
\bibliographystyle{agsm}
\bibliography{NNForecast.bib}

\newpage\phantom{}
\begin{appendices}
\section{Dataset}
\label{appendix:dataset}
\begin{enumerate}
    \item \texttt{JD-demand}. The \texttt{JD-demand} dataset is a collection of 50,000 time series of regional demand which involves around 6,000 products of \texttt{3C} (short for communication, computer and consumer electronics) category from seven regions of China. The dataset is gathered from \DTMdisplaydate{2014}{1}{1}{-1} to \DTMdisplaydate{2018}{12}{1}{-1}. 
        The features set for \texttt{JD-demand} includes historical demand and the product-specific information (e.g., region\_id, product categories, brand, the corresponding product price and promotions).
    \item \texttt{JD-shipment}. %Shipment forecasting means to forecast the delivery volume of daily packages for warehouses or distribution center, e.g., the daily number of packages of "shoe-A" warehouses of Beijing city for next month. The standard same-day delivery and next-day delivery services provided by JD.com result in huge pressure in logistics systems. Therefore, accurate forecasting is of extreme importance as it can help the company to do better logistics operations and reduce operations costs.
The \texttt{JD-shipment} dataset includes about 1450 time series from \DTMdisplaydate{2014}{10}{1}{-1} to \DTMdisplaydate{2018}{12}{1}{-1}, including new series (warehouses) that emerge with the development of the companies' business. The covariates consist of historical demand, the warehouse specific info including geographic and metropolitan informations (e.g., geo\_region, city) and warehouse categories (e.g. food, fashion, appliances).
    \item \texttt{Electricity}. The \texttt{electricity} dataset describes the series of the electricity consumption of 370 customers, which can be accessed at \url{https://archive.ics.uci.edu/ml/datasets/ElectricityLoadDiagrams20112014},  The electricity usage values are recorded per 15 minutes from 2011 to 2014. We select the data of the last three years. 
By aggregating the records of the same hour, we use the hourly consumption data of size $N \times T = 370 \times 26304$, where $N$ is the number of time series and $T$ is the length~\citep{yu2016temporal}.
    \item \texttt{Traffic}. The \texttt{traffic} dataset describes the occupancy rates (between 0 and 1) of 963 car lanes from San Francisco bay area freeways, which can be accessed at \url{https://archive.ics.uci.edu/ml/datasets/PEMS-SF} 
        The measurements are carried out over the period from \DTMdisplaydate{2008}{1}{1}{-1} to \DTMdisplaydate{2009}{03}{30}{-1} and are sampled every 10 minutes. 
The original dataset was split into training and test parts, and the daily order was shuffled.
The total datasets were merged and rearranged to make sure it followed the calendar order.
Hourly aggregation was applied to obtain hourly traffic data~\citep{yu2016temporal}. 
Finally, we get the dataset of size $N \times T = 963 \times 10560$, with the occupancy rates at each station described by a time series of length $10,560$. 
  \item \texttt{Parts}. The \texttt{parts} dataset includes 2,674 time series supplied by a US car company, which represents the monthly sales for slow-moving parts and covers a period of 51 months. The data can be accessed at  \url{http://www.exponentialsmoothing.net/supplements\#data}. After applying two filtering rules as follows:
        \begin{itemize}
            \item Removing series possessing fewer than ten positive monthly demands.
            \item Removing series having no positive demand in the first 15 and final 15 months.
        \end{itemize}
        There are finally 1,046 time series left and a more detailed description can be find in ~\cite{hyndman2008forecasting}.
\end{enumerate}

\section{Baselines}
\label{appendix:baseline}
    %\item \texttt{lightGBM}: Gradient boosting tree method has been empirically proven to be a highly effective approach in predictive modeling. As one of efficient implementation of the gradient boosting tree algorithm, \texttt{lightGBM} has gained popularity of being the winning algorithm in numerous machine learning competitions, like Kaggle Competition~\cite{chen2016lightGBM}.

Forecasting in industrial applications often relies on a combination of univariate forecasting models and machine-learning based methods.
\begin{enumerate}
    \item \texttt{SARIMA}: Seasonal ARIMA (\texttt{SARIMA}) is a widely used time series forecasting model which extends the ARIMA model by including additional seasonal term and is capable of modeling seasonal behaviors from the data~\citep{box2015time}. Currently, \texttt{SARIMA} is applied to \texttt{JD-shipment} dataset and fast-moving products with historical data of length  more than 14 in \texttt{JD-demand} dataset. 
The model is implemented with Python's package \texttt{pmdarima}~\citep{Taylor2017} and the best parameters are automatically select based on the criterion of minimizing the AICs~\citep{hyndman2018forecasting}. The predictions at confidence level \{10\%, 90\%\} are taken as the probabilistic forecasts in our experiments.
    \item \texttt{lightGBM}: Gradient boosting tree method has been empirically proven to be a highly effective approach in predictive modeling. 
        As one of efficient implementation of the gradient boosting tree algorithm, \texttt{lightGBM} has gained popularity of being the winning algorithm in numerous machine learning competitions, like Kaggle Competition~\citep{ke2017lightgbm}.
        \texttt{lightGBM} is applied to both \texttt{JD-demand} dataset and \texttt{JD-shipment} dataset. The features for forecasting on \texttt{JD-shipment} are presented in Table~\ref{table:xgb}.  
        A grid-search is used to find the best values of parameters like learning rate, the depth-of-tree based on the offline evaluation on data from both last month and the same month of last year. 
\end{enumerate}

\begin{table*}[!b]
            \caption{lightGBM feature lists}
    \label{table:xgb}
    \centering \small
    \begin{center}
    \begin{tabular}{ll}
    \toprule
        Feature type & Details \\ \midrule
        Category & region\_id, city\_id, warehouse\_type, holiday\_indicators,\\ 
         & is-weekend, etc.  \\ 
        Stats of warehouse level& summary (mean,median) of last week and last two weeks, \\
        & summary (median, standard deviation) of last four weeks, etc . \\ 
        Stats of city level& summary (mean,median) of last week and last two weeks, \\
        & summary (median, standard deviation) of last four weeks, etc .  \\ 
        Stats of warehouse-type level & summary (mean,median) of last week and last two weeks, \\
        & summary (median, standard deviation) of last four weeks, etc.  \\ \bottomrule
        %Warehouse-level  &median-of-last-week & median-of-last-two week & median-of-last-four-week & sd-of-last-four-week   &   \\ %\hline
        %warehouse\_type-level & median-of-last-week & sd-of-last-four-week & & & \\ 
        %city\_type-level & median-of-last-week & sd-of-last-four-week & & & \\  \bottomrule
       % best iteration &  & 60m &  225 & 167 &  11 \\ \bottomrule
    \end{tabular}
    \end{center}
\end{table*}

\begin{table*}[!b]
            \caption{Dataset details and deepTCN parameters}
    \label{table:training}
    \centering
    \begin{center} \small
    \setlength\tabcolsep{4pt}
    \begin{tabular}{lccccc}
    \toprule
       & \texttt{JD-demand} & \texttt{JD-shipment}   & \texttt{electricity-quantile} & \texttt{traffic} & \texttt{parts} \\ \midrule
        number of time series & 50,000 & 1,450 & 370  & 963   & 1,406\\ %\hline
       % time granularity &daily & daily & hourly & hourly &  monthly  \\%\hline
       % domain  & $\N$ &  $\N$ & $\R^{+}$ &  $[0,1]$ & $\N$\\ %\hline
        input-output length &31-31 & 31-31 & 168-24 & 168-24  & 12-12 \\ %\hline
        dilation-list& [1,2,4,8]  & [1,2,4,8]   &  [1,2,4,8,16,20,32] & [1,2,4,8,16,20,32]  & [1,2] \\ %\hline
        number of hidden layers  & 6 & 6 & 9  & 9   & 4\\ %\hline
        number of training samples & 200k & 40k  &  30k &  26k & 4k \\ %\hline
        batch size & 16 & 512 & 512  & 128 & 8 \\ %\hline
        learning rate &1e-2  & 5e-2 &5e-2   & 1e-2 &1e-4  \\ \bottomrule
       % best iteration &  & 60m &  225 & 167 &  11 \\ \bottomrule
    \end{tabular}
    \end{center}
\end{table*}

\section{Experiment details}
The current model is implemented with \texttt{Mxnet}~\citep{chen2015mxnet} and its new high-level interface Gluon. We trained our model on a \texttt{GPU} server with one \texttt{Tesla P40} and 16 CPU (3.4 \texttt{GHz}). Multiple-GPU can be applied to speed up and achieve better training efficiency in real industrial application. The code for public datasets is available from \url{https://github.com/oneday88/deepTCN}.

For the JD.com's datasets, the training range and prediction horizon are both 31 days. We implement two models for both \texttt{JD-demand} and \texttt{JD-shipment} datasets. One model is trained on the data before Oct 2018 and produces forecasting on Oct 2018; the other one is trained on the data before Nov 2018 and produces forecasting on Nov 2018.

For the \texttt{parts} dataset, we use the first 39 months as training data and the last 12 months for evaluation. 
A rolling window approach with window size =4 is adopted. 
The training and prediction range are both 12 months and a rolling window approach with window size 4 is adopted. For both \texttt{electricity} and \texttt{traffic} datasets, the training range and prediction range are selected as 168 hours and 24 hours respectively. For \texttt{electricity} dataset, we use only samples taken in December of 2011, 2012 and 2013 as training data, as we assume that this small data set is sufficient for the task of forecasting electricity consumption during the last seven days of December 2014. For \texttt{traffic} dataset, we train models on all the data before last seven days.

For each dataset, we fit the model on the training data and evaluate the corresponding metrics on the testing data after every epoch. When the training process is complete, we pick the model that gains the best evaluation results on the test set.

Convolution-related hyper-parameters, such as kernel size, number of channels and dilation length, are selected according to different tasks and datasets. The most important principle for choosing kernel size and dilation length is to make sure that the encoder (stacked residual blocks) has sufficiently large receptive field, namely long effective history of the time series. The number of channels at each convolution layer is determined by the number of input features and is kept fixed for all residual blocks. We manually tune for each dataset training-related hyper-parameters, including batch size and learning rate, in order to achieve the best performance on both evaluation metrics and running time. A more detailed description of parameters is presented in Table ~\ref{table:training}.

\end{appendices}

\end{document}